\documentclass[lettersize,journal]{IEEEtran}
\usepackage{amsmath,amsfonts}
\usepackage{algorithmic}
\usepackage{array}
\usepackage[caption=false,font=normalsize,labelfont=sf,textfont=sf]{subfig}
\usepackage{textcomp}
\usepackage{stfloats}
\usepackage{url}
\usepackage{verbatim}
\usepackage{graphicx}
\graphicspath{{images/}}  
\usepackage{balance}
\usepackage{comment}
\usepackage{multicol}
\usepackage{booktabs}
\usepackage{multirow}
\usepackage{listings}
\usepackage{xcolor}
\usepackage{hyperref}
\usepackage{cleveref}

\begin{document}

\title{Robotic surface exploration with vision and tactile\\
sensing for cracks detection and characterisation} 

\author{Francesca Palermo$^{1,2 *}$, Bukeikhan Omarali$^{1}$, Changae Oh$^{1}$, Kaspar Althoefer$^{1}$, Ildar Farkhatdinov$^{1,3 *}$ 
	\thanks{$^{1}$ The Centre for Advanced Robotics @ Queen Mary (ARQ), School of Electronic Engineering and Computer Science, Queen Mary University of London, London, United Kingdom {\tt\small }}%
    \thanks{$^{2}$ Department of Medicine, Imperial College of Science, Technology and Medicine; London, United Kingdom.{\tt\small }}%
	\thanks{$^{3}$ Department of Bioengineering, Imperial College of Science, Technology and Medicine; London, United Kingdom.{\tt\small }}%
	\thanks{$*$ \{f.palermo,i.farkhatdinov\}@qmul.ac.uk}
}


\maketitle

\begin{abstract}
This paper presents a novel algorithm for crack localisation and detection based on visual and tactile analysis via fibre-optics. 
A finger-shaped sensor based on fibre-optics is employed for the data acquisition to collect data for the analysis and the experiments.
To detect the possible locations of cracks a camera is used to scan an environment while running an object detection algorithm.
Once the crack is detected, a fully-connected graph is created from a skeletonised version of the crack. 
A minimum spanning tree is then employed for calculating the shortest path to explore the crack which is then used to develop the motion planner for the robotic manipulator.
The motion planner divides the crack into multiple nodes which are then explored individually.
Then, the manipulator starts the exploration and performs the tactile data classification to confirm if there is indeed a crack in that location or just a false positive from the vision algorithm. 
If a crack is detected, also the length, width, orientation and number of branches are calculated.
This is repeated until all the nodes of the crack are explored.
In order to validate the complete algorithm, various experiments are performed: comparison of exploration of cracks through full scan and motion planning algorithm, implementation of frequency-based features for crack classification and geometry analysis using a combination of vision and tactile data.
From the results of the experiments, it is shown that the proposed algorithm is able to detect cracks and improve the results obtained from vision to correctly classify cracks and their geometry with minimal cost thanks to the motion planning algorithm.
\end{abstract}


\section{Introduction}


\IEEEPARstart{D}{etecting} mechanical fractures on an object, such as containers and pipes, is an important task often performed in remote hazardous environments.
In this situation, crack detection is particularly important since it can avoid spillage of hazardous material from the container or detect cracks on the surface of the concrete.
Majority of existing crack detection approaches rely on computer vision techniques of the examined section~\cite{mohan2018crack}, eddy current implementation in metallic structures~\cite{yao2014crack}, or ultrasonic techniques~\cite{chang2017adaptive}.

In supervised environments in which the cracks have clear continuity and, when acquired with a camera, can produce high contrast images, edge detection and image segmentation methods can be implemented.
However, cracks are more commonly encountered in noisy backgrounds, resulting in poor continuity, low contrast, and a detrimental impact on the acquired picture quality.
Deep learning-based approaches for locating and classifying cracks have recently been developed~\cite{dorafshan2018comparison}.
Chen et al~\cite{chen2017nb} proposed a fusion deep learning framework called NB-CNN (Na\"{i}ve Bayes - Convolutional Neural Network) which discovered crack patches in each video frame by analysing individual video frames for crack detection. 
DeepCrack, a deep convolutional neural network for automatic crack identification, was proposed by Zou et al~\cite{zou2018deepcrack}.
It uses multi-scale deep convolutional features learned at hierarchical convolutional stages to recognise line formations.
VGG-16 DCNN, a pre-trained deep neural convolutional neural network, was used by Gopalakrishnan et al.~\cite{gopalakrishnan2017deep} to automatically detect fractures in two surface pavement datasets.

Images captured from a camera can be further analysed in the frequency domain applying wavelets.
Subirats et al.~\cite{subirats2006automation} presented a method for crack detection based on 2D continuous wavelet transform to create a binary image which indicates the presence or not of cracks on the pavement surface image.
Zhou et al.~\cite{zhou2006wavelet} proposed an algorithm to separate cracks on roads from noise and background using statistical criteria developed through wavelet coefficients. 
In~\cite{jiang2012crack} Jiang et al. introduced a method for detecting beams based on complex Continuous Wavelets Transforms which is more robust when the signal is contaminated by noise in respect to the simple Continuous Wavelets Transforms.
Considering the anisotropic characteristics of wavelet transformations, these techniques may be at disadvantage when analysing cracks with high curvature or reduced continuity.
The above-described crack detection methods are based on computer vision techniques and can fail in remote environments with limited luminosity or noise due to radiation. 
Furthermore, vision-based methods are not capable of acquiring material properties such as texture and hardness.

In contrast to the visual modality, tactile and proximity sensing can provide important information on material properties such as shape, texture and hardness~\cite{kappassov2015tactile,tomo2016design}.
The stiffness of objects has been investigated~\cite{konstantinova2017object} implementing a hybrid force and proximity finger-shaped sensor achieving 87\% classification accuracy on a set of household objects with different stiffness values.
In~\cite{palermo2020implementing, palermo2020automatic} it was demonstrated how to use fibre optics to recognise and classify fractures on surfaces using time-domain features.
Jiang et al.~\cite{jiang2021vision} proposed a vision-tactile algorithm for detecting cracks using RGB-D images segmented with fine-tuned Deep Convolutional Neural Networks and a set of contact points generated to guide the collection of tactile images by a camera-based optical tactile sensor. 
During contact between the sensor and the crack, a pixel-wise mask of the crack was obtained from tactile images to improve the shape of the crack.
In addition to machine learning algorithms which need engineered features extracted from the data, deep learning models (such as CNN) can be applied to tactile analysis when converting tactile data to their respective figures. 
The authors of~\cite{alameh2019dcnn} suggested an algorithm for recognising the object touched on an electronic skin via human interactions.
The skin's 3D tactile data was transformed into 2D pictures and fed into a CNN that outperformed traditional tactile data classification methods.
In~\cite{taunyazov2019towards}, the effects of combining touch and sliding movement for tactile texture categorisation via CNN were investigated.
It was shown that touch data can be used to make an initial estimate, which can then be revised via sliding.
The authors of~\cite{gao2016deep} demonstrated that using a multi-modal technique based on CNN with both visual and physical contact signals resulted in more accurate findings and more robust classification compared to hand-designed features methods.
\textbf{Proposed Approach.}
In this paper, as shown in Figure~\ref{fig:algorithm}, we propose a multi-modal algorithm based on computer vision and tactile analysis to detect and classify cracks. 
First, object detection is applied to detect possible cracks in a scene.
The extracted figures are then further analysed via a graph theory algorithm to create a motion planner for a robot manipulator with a tactile sensor attached as an end-effector.
Each of the detected cracks is then explored via the proposed motion algorithm and a machine learning classifier confirms if there is indeed a crack or just a false positive from vision.
For each of the detected crack additional geometry information are calculated.

\section{Proposed System}
\label{sec:proposed_system}
\begin{figure*}
    \centering
    \includegraphics{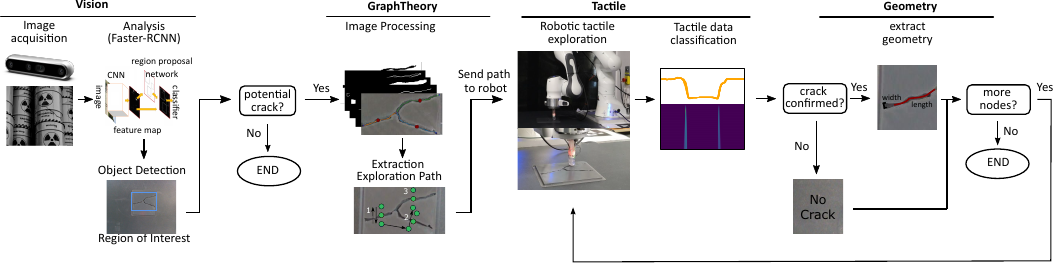}
    \caption{Complete Algorithm for the crack detection and exploration. 
    First, faster R-CNN is performed to detect the possible location of cracks. 
    Once a crack is detected, a graph theory algorithm is performed to extract the motion planning algorithm for the robotic manipulator The motion planning divides the crack into multiple nodes which are then explored one by one. 
    At this point, the manipulator starts the exploration and performs the tactile data classification to confirm if there is indeed a crack in that location or just a false positive from the vision algorithm. 
    This is repeated until all the nodes of the crack are explored.}
    \label{fig:algorithm}
\end{figure*}

For real-time applications, exploring whole surfaces using only a tactile approach would be too time-consuming and may produce errors.
Because of this, we propose a multi-modal algorithm based on a combination of vision and tactile modalities shown in Figure~\ref{fig:algorithm}. 
First, the camera scans the environment and faster R-CNN is performed to detect the possible location of cracks as described in \cite{palermo2021multi}.
Once the crack is detected, a graph theory algorithm is performed to extract the motion planning algorithm for the robotic manipulator, as described in Section~\ref{sec:graph_theory}.
The motion planning divides the crack into multiple nodes which are then explored individually.
Then, the manipulator starts the exploration and performs the tactile data classification to confirm if there is indeed a crack in that location or just a false positive from the vision algorithm, described in Section~\ref{sec:tactile_data_acquisition}.
This is repeated until all the nodes of the crack are explored.

The finger-shaped sensor based on fibre-optics described in~\cite{konstantinova2017object} was employed for the data acquisition to collect data for the analysis and the experiments.
The sensor is made of two 3D-printed rigid parts acting as the distal and proximal phalanges of a finger and one 3D-printed soft part, the intermediate phalanx, positioned among the two rigid phalanges.
Three pairs of fibre optics (D1, D2, D3) are used to measure the sensor's soft part deformation via changes in the reflected light intensity. 
A fourth pair of optical fibre cables (P) is positioned at the tip of the finger and it is used to sense the proximity to external objects.

The sensor was attached as an end-effector to a Franka Emika's Panda robot~\footnote{https://www.franka.de/} which was used to explore and acquire the data from surfaces of interest.
The experimental setup with the video camera and the tactile sensor mounted at the end-effector of a robotic manipulator is shown in Figure~\ref{fig:samples} a).
Different 3D-printed surfaces were provided for exploration.

\subsection{3D Printed Samples}
A set of 13 (9 different cracks geometry, 3 bumpy surfaces and 1 flat surface) 3D printed surfaces was employed for the experiments.
Each of the surfaces was 3D printed with an Ultimaker~III, 0.2~mm layer height, 0.4~mm nozzle diameter. 

\begin{figure}
    \centering
    \includegraphics[width=0.9\columnwidth]{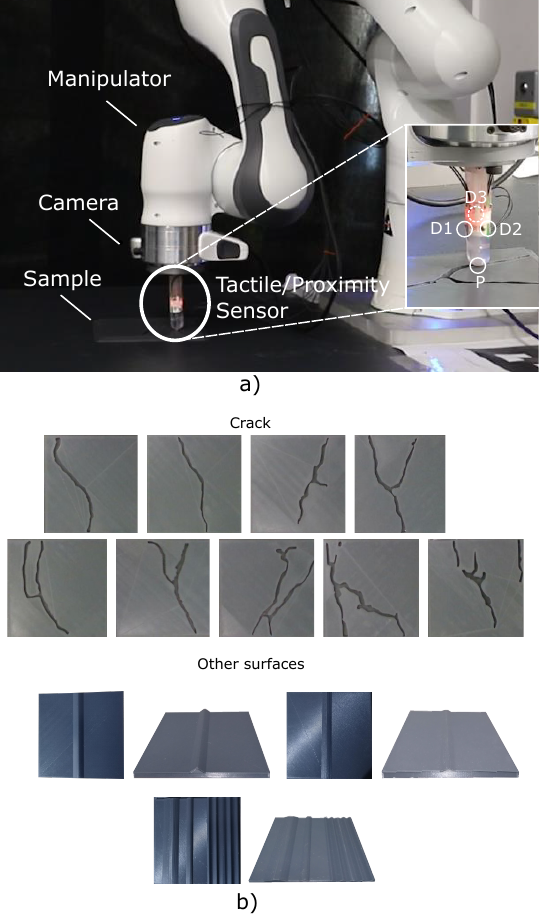}
    \caption{a) The experimental setup for the surface exploration multi-modal approach with vision and the tactile sensor consisting of 3 pairs of fibre optics to calculate the deformation of the soft middle part of the sensor (D1, D2 and D3) and a pair of fribre optics positioned at the tip to calculate the proximity value. b) 3D printed samples of real cracks and bumpy and flat surfaces: 3 simple cracks, 3 y-shaped cracks and 3 difficult cracks, 2 possible combinations of bumpy surfaces and a flat surface.}
    \label{fig:samples}
\end{figure}

The crack surfaces were extracted by analysing real crack images. 
Using Inkscape, it was possible to extract the bitmap of the images and to create a model in Blender which was then converted into an .stl file and 3D printed using the Ultimaker Cura software.
Each model is 125x125x5 mm in size but has a different shape, length, and width.
Furthermore, a flat surface was printed with the same size.
3 different bumpy surfaces are also printed to create more types of possible non-cracked surfaces.
A sample of the surfaces used for the acquisitions and the experiments is shown in Figure~\ref{fig:samples} b). %

\section{Motion Planning with Vision for Tactile Exploration}

\subsection{Experimental Methods for Motion Planning}
\label{sec:graph_theory}
\begin{figure*}
	\centering
	\includegraphics[width=0.9\textwidth]{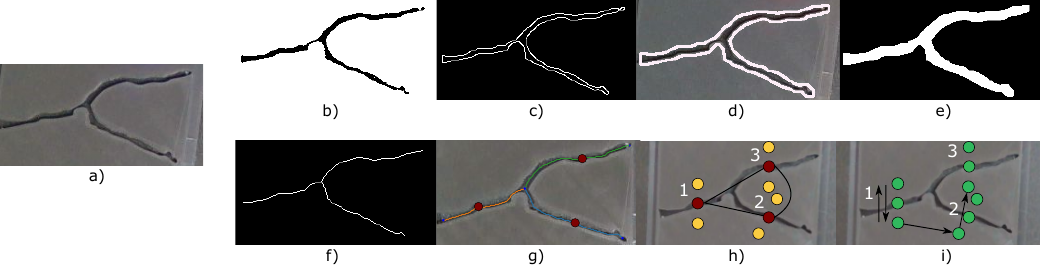}
	\caption[Example of image processing for the crack's geometry analysis]{Example of image processing for the crack's geometry analysis: (a) original image; 
		(b) binary image; 
		(c) threshold Canny edge detection with morphological transformations; 
		(d) extraction of contours; 
		(e) binary mask; 
		(f) pruned skeleton; 
		(g) graph and branches identified, in red the middle points of each edge are shown; 
		(h) the fully-connected middle points graph is created; 
		(i) the optimal exploration path is defined (only the node 1 left and right exploration are shown for brevity).}
	\label{fig:edge_detection}
\end{figure*}

Previously, a multi-modal robotic visual-tactile algorithm was developed to detect and localise possible fractures \cite{palermo2020automatic, palermo2021multi}.
The method employed Faster-RCNN for fracture detection.
Once the region of interest containing fractures was extracted, the images could be further explored by extracting the geometry information to plan an optimally controlled tactile exploration.

To analyse the obtained localised crack images, image processing and computer vision techniques were implemented to create a skeletonised version of the image of the fracture which was then transformed into a graph and explored via graph theory.
The key steps are demonstrated in Figure~\ref{fig:edge_detection}.
First, to avoid obtaining open contours and incomplete masks, a uniform coloured padding was introduced in the acquired image (size 100$\times$200 pixels) of the fracture (Figure~\ref{fig:edge_detection}a) to close potentially open locations.
The colour of the padding was chosen as the average of the total RGB colours of the image.
The original image was then converted to greyscale and blurred with a Gaussian filter (3x3 kernel).
The resulting image was converted to a binary image using Otsu and binary thresholds (Figure~\ref{fig:edge_detection}b).
To connect the disconnected cracks in the estimated binary image, the dilation operation in morphological transformations was applied. 
Then, Canny Edge detection was implemented (Figure~\ref{fig:edge_detection}c). 
The average of the intensities of the pixels was used to automatically estimate the lower and higher threshold for edge detection.
The resulting edges were improved with an additional dilation operation.
Using the obtained edges, the contours of the fractures were calculated (Figure~\ref{fig:edge_detection}d).
The averaged area of the contours was used to eliminate any outliers.
The object mask was then created (Figure~\ref{fig:edge_detection}e), which was used to skeletonise the fracture (Figure~\ref{fig:edge_detection}f).
For this purpose, we used an open-source PlantCV library~\cite{gehan2017plantcv} 
\footnote{\url{https://github.com/danforthcenter/plantcv}} which provides a useful method to create a skeleton from the mask and to prune it. 
The \textit{sknw} library\footnote{\url{https://github.com/Image-Py/sknw}} was applied to the resulting skeleton of the crack to convert it into a graph in which each ending point and branching point (a point in which various branches of the crack are created) were the vertices and the lines connecting these points were the edges (Figure~\ref{fig:edge_detection}g). 
This graph was further reduced by calculating the middle point of each of the branches (edges) and using those points as vertices of the new graph with \textit{Networkx} library\footnote{\url{https://github.com/networkx/networkx}} (Figure~\ref{fig:edge_detection}h).
In addition, for each middle point, shifted left and right points were created which were used to develop the motion plan of the manipulator with the fibre optic sensor, described in~\cite{palermo2020automatic}, attached as the end-effector.
These coordinates and weights were used to define the tactile exploration path. 
Being the robotic manipulator external to the surface, it does not have to comply with the curves of the crack to move from one point to the other. 
Thus, the edges can be converted to a straight line connecting the two vertices. 
The Euclidean distance between each vertex corresponds to the weight to move from one node to the other.
To find the least costly path to explore the whole graph, a revised version of the Minimum Spanning Tree was implemented.
In the proposed scenario, the explorable graph was bidirectional for each node and there was no specific starting point.
The main goal was to explore all the vertices only once with the minimum weight which corresponds to the minimum total Euclidean distance.
Each node was then analysed as starting point and all the possible paths were explored.
The node which produces the least expensive path based on the sum of the Euclidean distances of all the branches was then chosen as starting point and the path was sent to the manipulator (Figure~\ref{fig:edge_detection}i).

As a result, a robotic manipulator with a tactile sensor attached at the end-effector can directly explore only the main elements of complex cracks branch by branch following the paths identified with the help of geometrical analysis of the image of the cracks.

\subsection{Comparison Between Motion Planning and Complete Explorations}
\begin{figure}
	\centering
	\includegraphics[width=0.9\columnwidth]{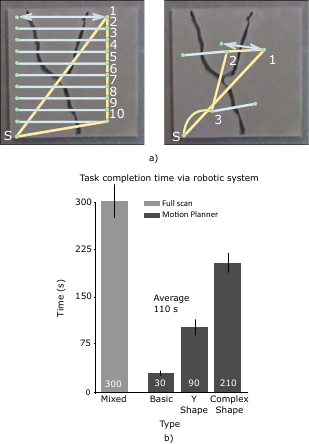}
	\caption[Example of crack exploration via motion planning]{a) On the left, it is shown the scanning movement of the manipulator. On the right, the motion planning obtained through the proposed graph theory algorithm is shown. For both images, yellow represents the movement of the robot to reach a specific point, in blue the movements corresponding to the node exploration and in green the start and end point for each exploration movement. Symbol S shows the starting position of the robot. For each blue movement, the robot is moving from one direction to the other and then backwards to the starting green point before moving to the next one. 
    b) Comparison of time required to explore cracked surfaces of various types (basic, y-shaped and complex-shaped) via full scans and motion planner during 10 runs. }
	\label{fig:ch3_scans}
\end{figure}

To validate the algorithm proposed in~\Cref{sec:graph_theory}, using the Franka Panda manipulator, the cracked surfaces shown in~\Cref{fig:samples}(b) were scanned for their whole entirety.
In total 9 cracked surfaces were employed: 3 simple-shaped cracks, 3 y-shaped cracks and 3 cracks of more complicated shape.
The surfaces were positioned in front of the robot and 10 points (from left to right) were sent to the control of the robot for the exploration.
This was repeated 10 times.
Figure~\ref{fig:ch3_scans} a) shows an example of scanning the whole surface on the left and of the implemented motion planning algorithm for crack exploration.

When scanning the entire surface, the scanning and classification took an average of $\sim$5 minutes, regardless of the crack's shape or number of branches.
When using graph theory to scan, the time required for exploration was determined by the number of branches and middle points found by the vision method.
The more branches there were, the longer it took to investigate the crack, simple ones with one or three branches required on average $\sim$30 seconds to two minutes to explore.
Cracks with a higher number of detected middle points required more time to be explored.
On average, it took $\sim$30 seconds for each branch of the fracture to be detected, analysed, and classified using the motion planning algorithm.
\Cref{fig:ch3_scans}(b) shows the comparison of the time required when exploring cracked surfaces via full scans and motion planner on the 9 cracks introduced in~\Cref{fig:samples}.
With the complete scanning to avoid losing possible minor cracks, it is necessary to accurately indicate the change in vertical exploration although this results in a longer exploration duration.
Using the proposed graph theory algorithm, it is possible to also detect the orientation of the crack and to perform the scanning perpendicularly to the crack. 
The complete scanning, on the other hand, has the risk of sliding parallel or over the crack's surface, missing the start and end of the branch of the crack and, consequently, information on the crack's width.
However, using the object detection algorithm and graph theory there is the possibility of not exploring cracks which are not detected by the vision algorithm. 
In conclusion, depending on the information required, scanning the entire surface of the crack may not be necessary; instead, only a few points extracted using graph theory can be used to obtain accurate information and speed up explorations.

\section{Tactile Detecting Cracks with Frequency Domain Features}
In the following section, it will be introduced the analysis of the tactile and proximity data acquired via the sensor described in~\ref{sec:proposed_system}. 
Machine learning techniques are implemented to classify the signals and features are investigated in the frequency domain to detect the presence of cracks on the explored surfaces.

\subsection{Tactile Data Acquisition}
\label{sec:tactile_data_acquisition}
Following the algorithm proposed in \Cref{fig:algorithm}, once a crack was detected by the object detection algorithm, introduced in \cite{palermo2021multi}, and further investigated via the graph algorithm proposed in \Cref{sec:graph_theory}, the motion planning for exploring the crack was sent to the Franka Panda controller via ROS Bridge. 
The experimental setup is shown in \Cref{fig:samples}(a). 
Two PCs communicated via ROS bridge with standard implementation on the ROS side and ROS-sharp implementation with Unity game engine that was used as a middleware for integration with Virtual Reality based user interface.
After the crack was detected and crack nodes were extracted, desired crack scan start and end positions in image pixel space were sent to the robot controller. 
Those pixel coordinates were then projected into the robot's 3D space resulting in a list of crack scan start and end positions. 
These 3D scan start and end positions were then added to the list of waypoints that would be used to plan the robot's trajectory. 
For each couple of scan start and end positions, the robot would begin a linear scanning motion at the start position until end position is reached, then reverse back to the start position. 
The robot would always approach to scan and retract from scan vertically. 
The finger was always oriented perpendicular to the scanned surface and the end-effector's y-axis was always parallel to the scan motion. 
The tactile data recording started when the robot finished the positioning and the approach to the scan start position and tactile data recording stopped when robot reached the end position of exploration and retracted from the scan. 
The tactile data then was used to classify the crack node and the robot would proceed scanning the remaining cracks. 

During the exploration of the crack, deformation and proximity data were recorded at 400~Hz via an Arduino Mega ADK micro-controller connected via serial port (USB) to the computer and further analysed via feature extraction.
To avoid any unwanted displacement during the movement, the samples were fixed to the laboratory desk during the acquisitions.
For each class (no crack, crack) $\sim$150 acquisitions were performed and each of the samples was differently positioned and oriented, \Cref{fig:samples}(b).

The data was first over-sampled from 400 Hz to 800 Hz and filtered with a Butterworth filter with a cutoff frequency of 30 Hz to extract frequency domain features.
The derivative was taken from these data and then filtered once more using a Butterworth filter.
The extracted features were then applied to the generated data.
Fourier transforms, spectrograms, continuous and discontinuous wavelets transformations, and other combinations of characteristics were investigated.

\subsection{Experimental Methods for Frequency Domain Features}
In previous works~\cite{palermo2020automatic, palermo2020implementing}, time domain features were investigated and analysed. 
The main concern with time domain features in this setup is that the implemented sensor highly relies on the position of the fibre optic cables and the colour of the explored surface.
In a previous version of the sensor, the cables were glued together with the 3D-printed parts of the sensor.
As a result, when one of the fibre optics cables broke, the whole sensor needed to be replaced.
To avoid having to replace the sensor and create additional waste, in the current version, the cables were left free.
Because of this, there was the possibility of one of the cables may slightly move and obtain different magnitudes in the data when relocating the sensor.
In the previous experiments, it was noticed that the implemented models with the time domain features were not robust to the movements of the cables when the sensor was moved from one physical location to another, even after applying standardisation and further optimisation techniques.

To overcome this problem, alternatives to the time domain features were investigated.
It was noticed that the shape of the data when exploring cracks was similar through different acquisitions performed in various experiments.
Frequency domain features were then explored to investigate the spectrum of the signal.

To extract frequency domain features, the data was first over-sampled from 400 Hz to 800 Hz and filtered with a Butterworth filter with a cutoff frequency of 30 Hz.
From this data, the derivative was extracted and filtered again with the Butterworth filter.
The resulting data was then used for feature extraction.
Multiple combinations of features were investigated: continuous and discrete wavelets transformations, Fourier transforms and spectrograms.

\subsubsection{Spectrograms}
From the above-mentioned data, to have a visual representation of the shape of the sensor and investigate the signal strength over time, the spectrograms of the derivative of the signal were extracted.
Spectrograms were normalised and the greyscale result was used.
The right side of~\Cref{fig:ch4_frequency_features} shows an example of obtained spectrogram for a flat surface, a bumpy surface and a surface with a crack for each of the fibre optic pairs of the implemented sensor.

\subsubsection{Fast Fourier Transform}
The Fast Fourier Transform~\cite{nussbaumer1981fast} (FFT) is a mathematical operation which converts a signal into individual spectral components and provides frequency information about the signal. 
FFTs have been used in multiple applications for quality control, and condition monitoring of machines or systems~\cite{muniategui2019one} and also for crack detection~\cite{abdel2003analysis}.
FFTs were calculated on the derivative of the signal data.
The left side of~\Cref{fig:ch4_frequency_features} shows an example of the FFT applied to the proximity P sensor for a flat surface, a bumpy surface and a surface with a crack.

\subsubsection{Discrete and Continuous Wavelets}
Although Fourier Transforms have a high resolution in the frequency domain, they have zero resolution in the time domain.
In order to investigate the problem of loss of components from the time domain, the use of Wavelet transform~\cite{rioul1991wavelets} has been proposed~\cite{sifuzzaman2009application}.
The Wavelet transform alters the shape of the Fourier transform's simple sine and cosine functions. 
In contrast to Fourier, where sine and cosine run from ($-\infty$,$+\infty$), the mother function in a Wavelet is finite in time. 
A wavelet decomposition, unlike a Fourier decomposition, uses a time-localised oscillatory function as the analysing or mother wavelet.
There are two possible types of Wavelets Transforms: Discrete and Continuous.
The difference between these two types of Wavelets Transforms is that the Continuous Wavelet Transform (CWT) uses an infinite number of scales and locations. 
On the other hand, the Discrete Wavelet Transform (DWT) makes use of a finite set of wavelets, which are defined at certain scales and locations.
Below, are the equations for Continuous Wavelets Transform:
\begin{equation}
	T(a, b)=\frac{1}{\sqrt{a}} \int_{-\infty}^{\infty} x(t) \psi^{*} \frac{(t-b)}{a} dt
\end{equation}
And Discrete Wavelet Transform:
\begin{equation}
	T_{m, n}=\int_{-\infty}^{\infty} x(t) \psi_{m, n}(t) d t
\end{equation}
Where $x$ is the original signal, $\psi$ is an arbitrary mother wavelet, $a$ is the scale factor, $b$ is the translation factor applied to the mother wavelet.
In addition to the two types of Wavelets, for each type, there are multiple possible families to choose from to best extract the informative data from the shape of the signal.
Discrete and Continuous Wavelets have been used in audio analysis~\cite{tzanetakis2001audio}, image processing~\cite{xizhi2008application} and crack detection~\cite{nigam2020crack}

\begin{figure*}
    \centering
    \includegraphics[width=0.8\textwidth]{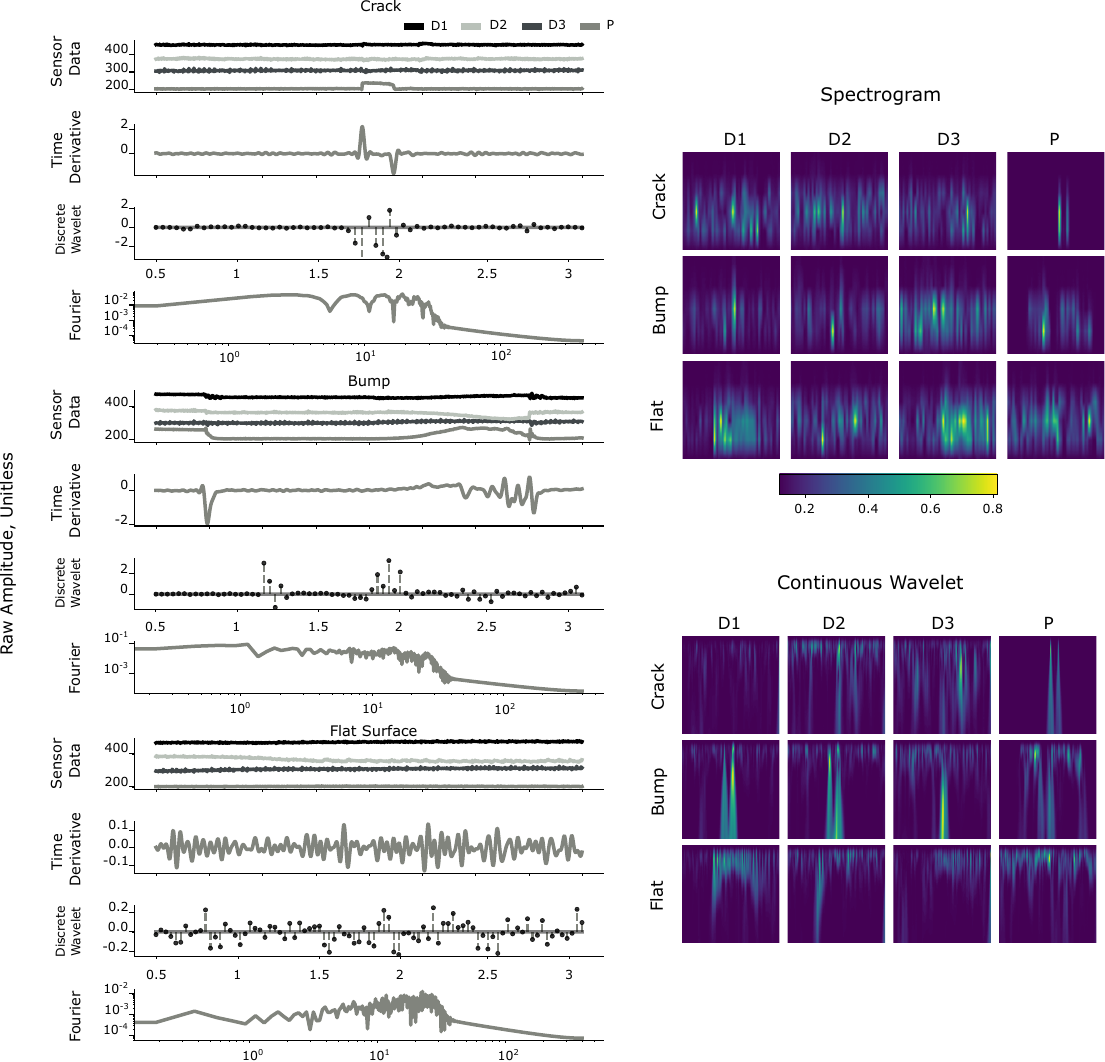}
    \caption[Engineered frequency domain features]{\textbf{On the left.} Sample the features implemented in the frequency data analysis for the proximity P sensor for crack, bumpy and flat surfaces. 
	For each surface, on the top the raw data acquired from the tactile sensor are shown, followed by the time derivative data extracted from the raw data.
	In the middle, the discrete wavelet transformations used in the Random Forest and Multi-modal fusion Convolutional Neural Network (M-CNN) models are shown.
	At the bottom, the Fourier Transformation implemented in the Random Forest and M-CNN models are shown.
	\textbf{On the right.} Sample of the features implemented in the frequency data analysis for crack, bumpy and flat surfaces. 
	For each surface, the spectrograms and continuous wavelets are shown that are implemented in the CNN and M-CNN models.
	}
    \label{fig:ch4_frequency_features}
\end{figure*}

In this work, the library Pywavelets~\cite{lee2019pywavelets} was used to investigate both types of Wavelets Transforms and all the corresponding families and to find the most accurate Wavelet Transform method for the implemented data.  
On the left of~\Cref{fig:ch4_frequency_features}, is shown an example of discrete wavelet transformation for flat, bumpy and cracked surfaces and on the right is an example of continuous wavelet transformation.

\subsection{Model for Crack Recognition with Tactile Sensing}
\label{sec:model}
For the analysis of the data, three models were implemented: Random Forest using fast Fourier transform and discrete wavelets, Convolutional Neural Network (CNN) for the spectrogram images and the continuous wavelets and Multi-CNN for all the above-mentioned features.

\subsubsection{Random Forest}
Random Forest~\cite{breiman2001random} is a machine learning algorithm which is used for both classification and regression tasks. 
It is an ensemble method which makes use of multiple learning trees.
It can handle missing data and it is more robust to over-fitting than other classifiers while producing at the same time reasonable predictions with little or no hyper-parameter optimisation.
Random Forest classifiers have been used for remote sensing and crack detection in multiple applications~\cite{belgiu2016random, palermo2020automatic, palermo2020implementing}.
In the following experiments, a Random Forest classifier with 100 trees has been implemented.

\subsubsection{Convolutional Neural Network}
\begin{figure*}
    \centering
    \includegraphics[width=\textwidth]{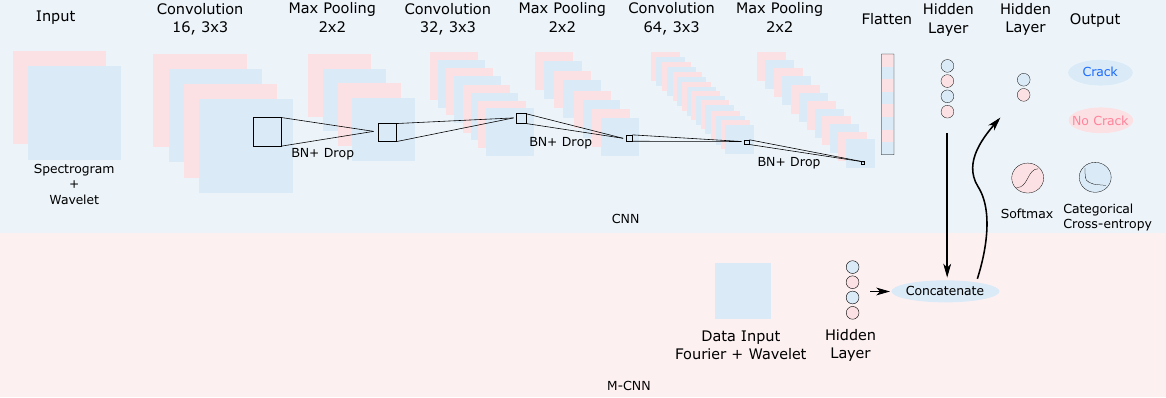}
    \caption[Developed CNN and M-CNN model architectures]{\textbf{CNN model architecture.} It uses a combination of the spectrogram and wavelet figure which consists of a 150x150x2 input shape. This is followed by a sequence which is repeated 3 times for a 16, 32 and 64 filter and consists of a Convolutional layer of 3x3 stride filter with Rectified Linear Unit (ReLU) activation, followed by batch normalisation and dropout layer, set to a 0.5 rate that helps prevent overfitting, and finally a Max Pooling layer of 2x2 stride filter. The final layer consists of a $softmax$ activated layer with two neurons.
	\textbf{M-CNN model architecture.} The data input consisting of Fourier and Wavelets extracted features is combined with the final layer of the CNN network.}
	\label{fig:architecture}
\end{figure*}

Convolutional Neural Network (CNN) is a type of Deep Learning algorithm made up of one or more convolutional layers.
It was first introduced in the 1980s with the "neocognitron" form for visual pattern recognition~\cite{fukushima1980self} and later evolved into the commonly known CNN and presented for handwritten character recognition~\cite{lecun1998gradient}.
CNN networks have been implemented in multiple applications for the detection of cracks~\cite{mohan2018crack}. 
In this work, a CNN model is implemented for classifying cracks using tactile and proximity data. 
Figure~\ref{fig:architecture} shows the complete architecture of the implemented CNN.
Based on the use of a spectrogram, the input of the model was the grayscaled image of the wavelet plus the grayscaled image of the spectrogram creating a 150x150x2 input shape.
The following sequence is then repeated thrice based on the filter size of the convolution layer which goes from 16 to 32 and 64:
Convolutional layer of 3x3 stride filter with Rectified Linear Unit (RELU) activation, followed by batch normalisation and dropout layer, set to a 0.5 rate that helps prevent overfitting, and finally a Max Pooling layer of 2x2 stride filter.
A \textit{softmax} with two neurons (the two possible values True or False for crack detection) was implemented as the output layer.
Categorical cross-entropy was used for loss and Adam~\cite{kingma2014adam} was used as an optimiser to minimise the cost function during training.
A simpler architecture with two Convolutional layers was also investigated but it resulted in higher overfitting.
The model is trained for 50 epochs to avoid overfitting.

\subsubsection{Multi Modal-Convolution Neural Network}
An alternative model is investigated to improve the results obtained with the best combination of features for both Random Forest and CNN with mixed data making use of TensorFlow Keras API.
Fast Fourier and the best Discrete Wavelet are used as numeric values while the spectrogram and the continuous wavelet are included as image data.
The Multi-modal fusion-Convolutional Neural Network (M-CNN) used the same architecture as CNN shown in Figure~\ref{fig:architecture} to analyse the figure's features.
Two hidden layers are used to analyse the discrete wavelets and Fourier transforms.
The two models were then concatenated.
Adam was used as optimiser and Categorical Crossentropy as loss.
The model was trained for 50 epochs.

\subsection{Experiments and Results}
To validate the developed models, introduced in Section~\ref{sec:model}, two experiments were performed. 
The first experiment used the dataset acquired with the Franka Panda robot as training and validation dataset to classify the data acquired in~\cite{palermo2020automatic} which were used as testing set.
The third experiment was performed to detect the crack online during the exploration with the Franka Panda manipulator.
In the experiment, the data was split 60\% for training, 20\% for validation and 20\% for testing. 

When using the Random Forest as the classifier, the discrete wavelets transform was used in combination with the Fast Fourier transformation.
It was investigated which one among discrete wavelets transformation or the Fast Fourier transformation was better for classifying the crack or if a combination of the two could improve the results.
For each feature, the number of peaks, the max value of the peak and the minimum value of the peak were extracted.

When using CNN and M-CNN as classifiers, the continuous wavelets transformation was implemented as features in combination with the Spectrograms of the data.
Following the same format as with the Random Forest, it was investigated if each feature would perform better than the other or if a combination of the two would be more accurate in detecting a cracked surface.
For brevity, in the following sections, the results with the wavelets achieving the best results are shown.
Furthermore, the models trained with Fourier and discrete wavelets or spectrogram and continuous wavelets are shown. 
In the experiment, a passive tactile exploration is performed by exploring the points extracted from the vision side of the algorithm. 
Thanks to this, it is possible to create a model of the crack and speed up its exploration. 
Active exploration could also be performed by sliding over the whole surface but it would incur in higher exploration time.

\subsubsection{Training and Testing with Different Database}
To investigate the ability to generalise of the models, an experiment was performed. 
The data acquired in~\Cref{sec:tactile_data_acquisition} were used as training set and the data acquired in~\cite{palermo2020automatic, palermo2020implementing} were implemented as testing set.
\Cref{tb:offline_true} shows the results of the classification.
\begin{table*}[]
	\centering
	\caption[Results for offline analysis via frequency domain features on data acquired with a different setup]{Results for offline analysis on data acquired with a different setup. D1,2,3 = Deformation data, P = Proximity data, RF = Random Forest, CNN = Convolutional Neural Network, M-CNN = Multi-modal fusion-CNN}
	\label{tb:offline_true}
	\resizebox{\textwidth}{!}{\begin{tabular}{cccc|lcccc}
		\toprule
		\multicolumn{1}{c}{\textbf{Model}} & \multicolumn{1}{c}{\textbf{Wavelet}}  & \multicolumn{1}{c}{\textbf{Fourier}} & \multicolumn{1}{c|}{\textbf{Spectrogram}} & \textbf{Sensor}     & \textbf{Accuracy} & \textbf{Precision} & \textbf{Recall} & \textbf{F1}  \\
		\midrule
		\multirow{6}{*}{RF} & \multirow{6}{*}{DB-11} & \multirow{6}{*}{T} & \multirow{6}{*}{F} & P & 88.33 $\pm$ 0.91 & 82.88 $\pm$ 1.21 & 99.6 $\pm$ 0.2 & 88.02 $\pm$ 0.99 \\
		&  &  &  & D1,2,3 & 55.5 $\pm$ 0.21 & 55.54 $\pm$ 0.1 & 99.8 $\pm$ 0.24 & 39.76 $\pm$ 0.28 \\
		&  &  &  & D1,2 + P & 88.33 $\pm$ 0.79 & 82.99 $\pm$ 1.13 & 99.4 $\pm$ 0.37 & 88.03 $\pm$ 0.86 \\
		&  &  &  & D1,3 + P & 88.33 $\pm$ 0.95 & 83.0 $\pm$ 1.35 & 99.4 $\pm$ 0.37 & 88.03 $\pm$ 1.03 \\
		&  &  &  & D2,3 + P & \textbf{88.56 $\pm$ 0.83} & \textbf{83.39 $\pm$ 1.28} & \textbf{99.2 $\pm$ 0.51} & \textbf{88.27 $\pm$ 0.91} \\
		&  &  &  & D1,2,3 + P & 88.44 $\pm$ 0.91 & 83.3 $\pm$ 1.21 & 99.1 $\pm$ 0.58 & 88.16 $\pm$ 0.98 \\
		\midrule
		\multirow{6}{*}{CNN} & \multirow{6}{*}{CGAU-1} & \multirow{6}{*}{F} & \multirow{6}{*}{T} & P & 78.78 $\pm$ 17.45 & 67.6 $\pm$ 34.15 & 76.4 $\pm$ 38.37 & 75.1 $\pm$ 24.13 \\
		&  &  &  & D1,2,3 & 52.06 $\pm$ 2.57 & 56.69 $\pm$ 2.88 & 67.7 $\pm$ 29.93 & 44.66 $\pm$ 3.0 \\
		&  &  &  & D1,2 + P & 73.94 $\pm$ 7.24 & 74.77 $\pm$ 3.4 & 82.5 $\pm$ 24.28 & 72.49 $\pm$ 7.67 \\
		&  &  &  & D1,3 + P & 76.5 $\pm$ 5.3 & 77.95 $\pm$ 4.53 & 81.3 $\pm$ 11.6 & 76.12 $\pm$ 5.49 \\
		&  &  &  & D2,3 + P & 77.56 $\pm$ 3.8 & 78.4 $\pm$ 4.02 & 83.3 $\pm$ 11.29 & 77.2 $\pm$ 3.72 \\
		&  &  &  & D1,2,3 + P & 71.78 $\pm$ 9.78 & 77.67 $\pm$ 10.41 & 77.6 $\pm$ 29.8 & 68.71 $\pm$ 12.5 \\
		\midrule
		\multirow{6}{*}{M-CNN} & \multirow{6}{*}{CGAU-1, DB-11} & \multirow{6}{*}{T} & \multirow{6}{*}{T} & P & 65.44 $\pm$ 11.98 & 81.2 $\pm$ 10.63 & 47.1 $\pm$ 27.16 & 62.02 $\pm$ 16.2 \\
		&  &  &  & D1,2,3 & 56.17 $\pm$ 1.96 & 57.07 $\pm$ 2.6 & 89.3 $\pm$ 10.39 & 47.03 $\pm$ 6.2 \\
		&  &  &  & D1,2 + P & 78.89 $\pm$ 4.24 & 80.84 $\pm$ 5.86 & 82.8 $\pm$ 10.56 & 78.53 $\pm$ 4.48 \\
		&  &  &  & D1,3 + P & 74.83 $\pm$ 11.04 & 83.5 $\pm$ 5.84 & 70.3 $\pm$ 25.95 & 73.24 $\pm$ 13.98 \\
		&  &  &  & D2,3 + P & 79.17 $\pm$ 5.33 & 76.84 $\pm$ 3.94 & 89.7 $\pm$ 8.55 & 78.66 $\pm$ 5.4 \\
		&  &  &  & D1,2,3 + P & 80.39 $\pm$ 5.73 & 83.54 $\pm$ 2.86 & 80.9 $\pm$ 13.09 & 80.25 $\pm$ 5.77
	\end{tabular}}
\end{table*}
The Random Forest is the most balanced one for almost all the various sensors combinations when using the proximity P and the deformations ones but it achieves the highest F1 score of $88.27\%$, recall $99.2\%$ and precision of $83.39\%$  when using a combination of the discrete wavelet transform DB-11 and the Fast Fourier Transform acquired from D2, D3 and P.
The CNN model achieves the highest score when considering a combination of proximity P for the deformation sensors D2 and D3, with an F1 score of $77.20\%$, recall of $83.33\%$ and precision of $78.40\%$ for the CGAU-1 continuous wavelet and the spectrogram.
For the M-CNN model, the best result was achieved when considering a combination of Fourier, Spectrograms, Continuous and Discrete Wavelets with an F1 score of $80.25\%$, recall of $80.90\%$ and precision of $83.54\%$ for D1, D2, D3 and P sensors, CGAU-1 as continuous wavelet and DB-11 as discrete wavelet.
Using the M-CNN model the sensitivity can be increased when considering only D2, D3 and P sensors at the expense of specificity.
Among the three models, the Random Forest Algorithm with the combination of implemented frequency-based features of Discrete Wavelet and Fourier Transformation is the most balanced and reliable model with fewer fluctuations among the various tactile data.

\subsubsection{Online Experiment with Different Modalities}
In addition to the offline experiment, one experiment was conducted online with three different modalities (single and multiple cracks, occluded cracks and painted cracks) by integrating the computer vision algorithm introduced in Section~\ref{sec:graph_theory} and the tactile analysis to further evaluate the generated classification models' abilities.
\begin{figure}
	\centering
	\includegraphics[width=\columnwidth]{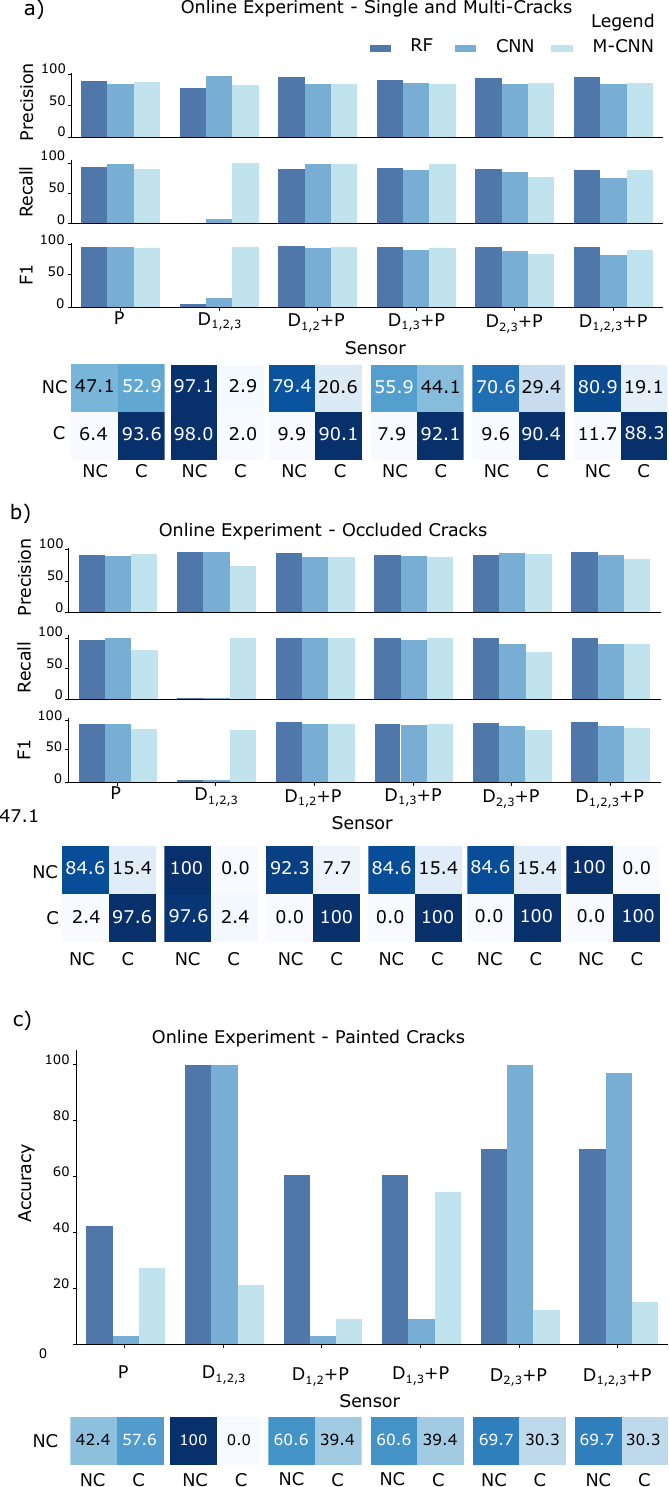}
	\caption[Results of online experiments for frequency domain analysis]{Results of the online experiment. For each modality, the results are compared when implementing the random forest classifier with the DB discrete wavelet transformation and Fourier transformation, the CNN with the CGAU continuous wavelet and spectrogram plot and the M-CNN with CGAU continuous wavelet and DB discrete wavelet transformations, Fourier transformation and spectrogram plots.
		NC = No Crack, C = Crack.\\
		(a) Modality 1 shows the results for online detection of single and multiple cracks.
		(b) Modality 2 shows the results of occluded cracks with screws, textile and other materials.
		(c) Modality 3 shows the results of fake cracks painted with a marker on surfaces.}
	\label{fig:ch4_frequency_online}
\end{figure}

\textbf{Online Experiment Modality 1 - Single and Multiple Cracks Detection.}
\label{sec:online_ex1}
First, each of the 3D printed samples introduced in Section~\ref{sec:graph_theory} was positioned in different orientations and explored 10 times for a total of 90 explorations and a total of 347 explored nodes.
Furthermore, multiple cracks were introduced into the environment and detected by the computer vision algorithm and then classified with the sensor. 
10 explorations were performed. 
For each exploration, each crack was divided into nodes which are classified one by one for a total of 64 nodes.
In total, 100 explorations were performed and 411 nodes were explored.
The models introduced in Section~\ref{sec:model} were used to classify the various samples.
In our setup, the weight of not detecting cracks present on the surface was higher than wrongly classifying a flat surface as crack. 
Undetected cracks may continue to grow and impact the functionality of the surface on which they are present.
Because of this, lowering the number of false negatives is crucial, but we also want to keep the number of false positives to a minimum.
Recall is thus the most crucial parameter in these studies, but F1, which takes both precision and recall into consideration, should also be taken into account.
\Cref{fig:ch4_frequency_online} (a) shows the results compared when implementing the random forest classifier with DB-11 discrete wavelet transformation and Fourier transformation, the CNN with CGAU-1 continuous wavelet and spectrogram plot and the M-CNN with CGAU-1 continuous wavelet and DB-11 discrete wavelet transformations, Fourier transformation and spectrogram plots.
The model which achieves the highest F1 score is the Random Forest classifier with 93.50$\%$, recall of 89.56$\%$ and precision of 97.90$\%$ using D1, D2 and P as sensors.
On the other hand, the model which achieves the highest recall of 99.30$\%$ at the expense of 86.51 $\%$ of precision is the M-CNN with D1,D2,D3 and P as sensors for a total of 92.48$\%$ F1 score.

\begin{figure}
	\centering
	\includegraphics[]{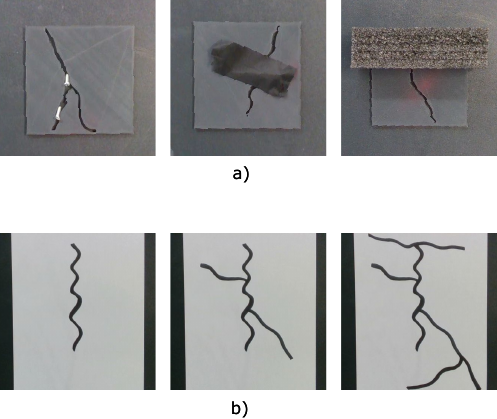}
	\caption[Example of occlusion and fake cracks]{(a) Example of the occlusion effect produced on the 3D printed cracked surfaces. 
		(b) Example of the cracks drawn with a marker.}
	\label{fig:sample_experiments}
\end{figure}

\textbf{Online Experiment Modality 2 - Detection of Occluded Crack.}
Occlusions were added to the cracks. 
The occlusions were created with screws and textile fabric. 
Figure~\ref{fig:sample_experiments} (a) shows an example of the produced occlusions. 10 explorations were performed for a total of 54 explored nodes.
Figure~\ref{fig:ch4_frequency_online} (b) shows the results compared when implementing the random forest classifier with DB-11 discrete wavelet transformation and Fourier transformation, the CNN with CGAU-1 continuous wavelet and spectrogram plot and the M-CNN with CGAU-1 continuous wavelet and DB-11 discrete wavelet transformations, Fourier transformation and spectrogram plots.
The model which achieves the highest F1 score is the random forest with 100$\%$, recall of 100$\%$ and precision of 100$\%$ using D1, D2, D3 and P as sensors.
When using the sensors D1, D2 and P, the model which achieves the highest result is still the random forest with 100$\%$ recall, precision of 97.62$\%$ and F1 score of 98.80$\%$.

\textbf{Online Experiment Modality 3 - Painted Crack.}
To test the robustness of the complete algorithm, fake cracks were created with a marker. 
Figure~\ref{fig:sample_experiments} (b) shows an example of the crack painted with the marker. 10 explorations were performed for a total of 33 explored nodes.
Figure~\ref{fig:ch4_frequency_online} (c) shows the results compared when implementing the random forest classifier with DB-11 discrete wavelet transformation and Fourier transformation, the CNN with CGAU-1 continuous wavelet and spectrogram plot and the M-CNN with CGAU-1 continuous wavelet and DB-11 discrete wavelet transformations, Fourier transformation and spectrogram plots.
Since this experiment was made up of negative labels only, the accuracy metric was used instead.
3 possible combinations of models achieved an accuracy score of 100$\%$: CNN or random forest with D1, D2 and D3 (only force sensors) and CNN with D2, D3 and P.
Introducing the proximity sensor P in the models resulted in higher false positive labels detected when classifying the data achieved from painted cracks. 
This may be due to the fact that when acquiring the data used for training the models, no similar data was added. 
Thus, when the model notices spikes in the proximity P was most prone to classify the data as a cracked surface.
On the other hand, when using only the force sensors, the data was classified as no crack because it was similar to a flat surface since no deformation in the sensors was detected.

\section{Crack characterisation via geometrical inspection}
In this section, a characterisation of the explored cracks is performed.
Using the data acquired in the previous sections, when a crack is detected, geometrical data are calculated for its length, width, orientation and number of branches.

\subsection{Crack Characterisation}
\label{ch5_crack_characterisation}
During the crack exploration, in addition to detecting the crack presence, further information characterising the crack can be extracted.
If a crack was detected during the exploration, the width, length, number of branches and orientation were calculated.
The set of cracked surfaces explored is shown in~\Cref{fig:samples}(b).

The width of the crack was determined by using the derivative of the fibre optics proximity data (P in Figure~\ref{fig:samples} (a)). 
From the derivative, the indexes of the highest slopes were extracted. 
These represent the start and end points of the width of the crack. 
The time necessary to explore this segment was then multiplied by the speed of the Franka Panda robot and the width was obtained.
The same was applied to both movements on the same section (from the start node to the endpoint and backward) and then the mean was extracted and used as the final width.
\begin{equation}
	\begin{split}
		T &= \frac{1}{f_s}\\
		t &= N T \\
		\overline{w} &= \frac{v_s  t_s +  v_e  t_e}{2}
	\end{split}
\end{equation}
where $T$ is the period, $f_s$ the sampling rate, $t$ the time necessary for exploring the section which is made of $N$ values, $v_s$ and $t_s$ indicate the velocity and time for the movement from start to end point, $v_e$ and $t_e$ the velocity and time for the opposite movement.
The length of the branches was calculated using the graph theory information. 
The x and y coordinates of the points of the branch were stored and used to calculate the length $l$ via the Euclidean distance:
\begin{equation}
	l =\sqrt{\sum_{i=0}^{n-1} {\left(x_{i+1} - x_{i}\right)^2} + \left(y_{i+1} - y_{i}\right)^2}
\end{equation}
The orientation of the branches was calculated using the start point of the branch and the endpoint and the arctangent between the two points.
\begin{equation}
	\theta = tan^{-1}(y_{e} - y_{s}, x_{e} - x_{s})  
\end{equation}
where $(x_{e}, y_{e})$ are the coordinates of the end point of the branch and $(x_{s}, y_{s})$ are the coordinates of the starting point of the branch.
The total number of branches of the crack was also calculated. 
When the classifier identified an explored segment as a crack, the number of branches was increased by 1.

\subsection{Experiments and Results}
\label{geometry_results}
\begin{figure}
	\centering
	\includegraphics[width=0.95\columnwidth]{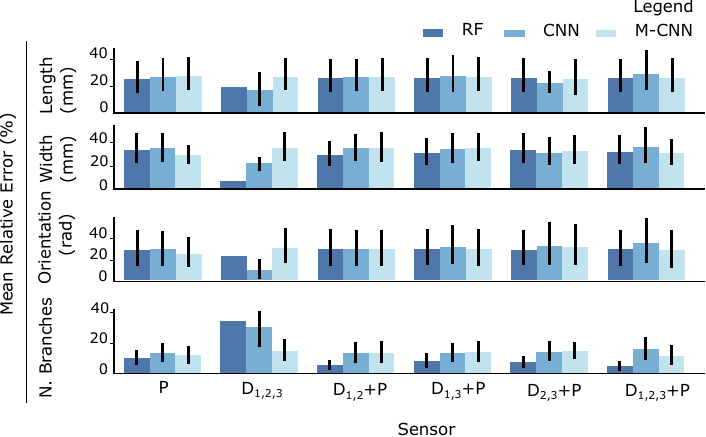}
	\caption[Results of the geometry experiments]{Results of the geometry experiments. Mean Relative Error is shown for length, width, orientation and number of branches. Width and Length are calculated in millimetres. Orientation is in degrees and the number of branches is an integer value.}
	\label{fig:ch5_geometry_results}
\end{figure}

Using the data acquired during Experiment 1 in Section~\ref{sec:online_ex1}, the geometry analysis of the fractures was performed.
Each of the explored node results was compared with the ground truths which were automatically created using an image of the fracture in plain visibility and with no additional external condition (e.g. occlusion, rotation, etc).
The Mean Relative Error (MRE) was calculated for length, orientation, number of branches and width.
Figure~\ref{fig:ch5_geometry_results} shows the results for the geometry experiments and the MAE error for the calculated measurements. 
Using only the deformations data for the Random Forest achieved the worst results since a majority of the cracks were mislabelled.
Thus, this model should not be taken into consideration in the following discussion.
The length measurements were calculated in millimetres. 
The maximum value for length was 157.13mm and the minimum value was 10.99mm.
The model showing the lowest MRE score of $\sim$18\% is the CNN with the deformations data.
The width measurements were calculated in millimetres. 
The biggest value was equal to 7mm and the minimum value was equal to 1.5mm.
The main problem with the width was that as the 3D printed samples were reconstructed from real cracks, the width changed from point to point and, as the cracks occurred in different orientation and positions, the middle point of the crack did not always correspond to the middle point of the ground truth and variations in millimetres occurred.
Nevertheless, the models achieved a MRE error of $\sim$30\%. 
The best model was the CNN using only the deformation data, achieving a $\sim$20\% MRE.
The orientation measurements were calculated in radians. 
In this case, the main concern was the different orientations of the captured pictures in respect to the ground labels of the original pictures. 
The majority of the acquisitions for experiments was a series of rotations of the 3d printed surfaces. 
Thus, by knowing the performed rotation, it was possible to remap the detected nodes with the ground nodes to appropriately calculate the degrees.
The best model for this measurement was the CNN with the deformation values which achieved a MRE of $\sim$15\%.
The number of branches measurement identifies the number of calculated sections of cracks. 
The maximum number of detected branches is 6 and the minimum is 1.
For this experiment, the best model was the Random Forest with a MRE of $\sim$9\% when using the deformation data together with the proximity data.

\section{Conclusions}
In this paper, an algorithm to detect and classify cracks is presented via visual and tactile data.
The method uses the camera to scan an environment and faster R-CNN is performed to detect the possible location of cracks.
Once a crack is detected, a graph theory algorithm is performed to calculate the least expensive motion planning sequence for the robotic manipulator.
The motion planning divides the crack into multiple nodes which are then explored individually.
Then, the manipulator starts the exploration and performs the tactile data classification to confirm if there is indeed a crack in that location or just a false positive from the vision algorithm.
If a crack is detected, also the length, width, orientation and number of branches are calculated.
This is repeated until all the nodes of the crack are explored.

In order to validate the complete algorithm, various experiments are performed.
From the results of the experiments, it is shown that the proposed algorithm is able to detect cracks and improve the results obtained from vision to correctly classify cracks and their geometry with minimal weight thanks to the motion planning algorithm.
This approach may be implemented also in extreme environments since gamma radiation does not interfere with the sensing mechanism of fibre optic-based sensors.

The paper has contributed to advances in crack detection by introducing a multi-modal algorithm that is used to detect cracks in the environment via computer vision and then confirming the presence of a crack via tactile exploration and machine learning classification of the data acquired from a fibre-optic-based sensor.
Few methods currently use tactile sensing for crack characterisation and detection and this is the first study which shows the reliability of tactile-based methodologies for crack detection via machine learning analysis. 
Furthermore, this is the first method which combines both tactile and vision for crack analysis. 

\textbf{Future Work.} 
The proposed algorithm was developed on flat surfaces and the camera was always perpendicular to the surface.
Considering the small region of interest corresponding to the detected crack for the exploration, the model may still perform with less accuracy. 
Proximity only may be used in this case to overcome displacement errors.
In the future, this may be improved by creating a depth mask of the explored surface through an RGB-D camera.

\bibliographystyle{IEEEtran}
\bibliography{main}

\end{document}